\documentclass[conference]{IEEEtran}
\IEEEoverridecommandlockouts

\usepackage[numbers,sort&compress]{natbib}
\usepackage{amsmath,amssymb,amsfonts}
\usepackage{algorithmic}
\usepackage{graphicx}
\usepackage{textcomp}
\usepackage{xcolor}
\usepackage{booktabs}
\usepackage{multirow}
\usepackage{url}
\usepackage{tikz}
\usepackage{hyperref}
\hypersetup{
    colorlinks   = true,
    citecolor    = blue,
    urlcolor    =  blue
}
\usetikzlibrary{arrows.meta, positioning, shapes.geometric}
\def\BibTeX{{\rm B\kern .05em{\sc i\kern .025em b}\kern .08em
    T\kern .1667em\lower.7ex\hbox{E}\kern .125emX}}

\definecolor{BluePrimary}{RGB}{31,78,121}
\definecolor{BlueSecondary}{RGB}{68,114,196}
\definecolor{GrayDark}{RGB}{64,64,64}
\definecolor{GrayMedium}{RGB}{160,160,160}
\definecolor{GrayLight}{RGB}{242,242,242}
\definecolor{OrangeAccent}{RGB}{237,125,49}
\definecolor{GreenAccent}{RGB}{112,173,71}
\definecolor{PurpleAccent}{RGB}{128,100,162}

\tikzset{
  headerbox/.style={
    rectangle,
    rounded corners=4pt,
    draw=BluePrimary,
    fill=BluePrimary!12,
    line width=1pt,
    text=GrayDark,
    minimum height=0.95cm,
    align=center,
    font=\small\bfseries\sffamily
  },
  bluebox/.style={
    rectangle,
    rounded corners=4pt,
    draw=BlueSecondary,
    fill=BlueSecondary!10,
    line width=0.9pt,
    text=GrayDark,
    minimum height=0.9cm,
    align=center,
    font=\small\sffamily
  },
  blueboxstrong/.style={
    rectangle,
    rounded corners=4pt,
    draw=BluePrimary,
    fill=BluePrimary!18,
    line width=1pt,
    text=GrayDark,
    minimum height=0.9cm,
    align=center,
    font=\small\bfseries\sffamily
  },
  annot/.style={
    font=\footnotesize\sffamily,
    text=GrayDark
  }
}

\makeatletter
\renewcommand*{\footnoterule}{%
  \kern -3pt                % Move up slightly
  \hrule width 1in height 0.4pt  % Make it 2 inches wide and thin
  \kern 2.6pt               % Move down
}
\makeatother

\begin{document}

\title{From Sinhala to Dhivehi: Cross-Lingual Transfer Learning for Low-Resource Speech Recognition}
% \thanks{Code available at: \url{https://github.com/lukmalilyas/FYP}}}

% \author{
%     \IEEEauthorblockN{Lukmal Ilyas}
%     \IEEEauthorblockA{\textit{School of Computing} \\
%     \textit{Informatics Institute of Technology}\\
%     Sri Lanka \\
%     ilyas.20221732@iit.ac.lk}
%     \and
%     \IEEEauthorblockN{Nevidu Jayatilleke}
%     \IEEEauthorblockA{\textit{Department of Computer Science \& Engineering} \\
%     \textit{University of Moratuwa}\\
%     Sri Lanka\\
%     nevidu.25@cse.mrt.ac.lk}
% }

\author{\IEEEauthorblockN{Lukmal Ilyas\IEEEauthorrefmark{1}, Nevidu Jayatilleke\IEEEauthorrefmark{2}}
\IEEEauthorblockA{\IEEEauthorrefmark{1}School of Computing, Informatics Institute of Technology, Sri Lanka\\
\texttt{ilyas.20221732@iit.ac.lk}}
\IEEEauthorblockA{\IEEEauthorrefmark{2}Department of Computer Science \& Engineering, University of Moratuwa, Sri Lanka\\
\texttt{nevidu.25@cse.mrt.ac.lk}}}

% \author{
%     Lukmal Ilyas$^a$ and Nevidu Jayatilleke$^b$ \\
%     $^a$School of Computing, Informatics Institute of Technology, Sri Lanka \\
%     $^b$Department of Computer Science \& Engineering, University of Moratuwa, Sri Lanka \\
%     \texttt{ilyas.20221732@iit.ac.lk, nevidu.25@cse.mrt.ac.lk}
% }

\maketitle

\begin{abstract}

Dhivehi, the national language of the Maldives, is currently under-resourced for automatic speech recognition (ASR) and other NLP tasks. This study investigates whether cross-lingual transfer learning from Sinhala, a linguistically related, relatively well-resourced Insular Indo-Aryan language, can improve Dhivehi ASR. We conduct seventeen experiments across five transfer learning paradigms: Dhivehi-only baselines, sequential fine-tuning, multilingual fine-tuning, continual pre-training, and a control using Turkish as an unrelated language. The strongest system, continual pre-training on Sinhala followed by fine-tuning on Dhivehi with \texttt{KenLM}, achieves 12.89\% WER and 2.70\% CER, outperforming the Dhivehi-only baseline by 13.50\% WER and 3.02\% CER. However, the adaptation strategy and decoding configuration are equally critical for a successful transfer learning experiment. We conduct seventeen controlled experiments spanning five transfer learning paradigms: Dhivehi-only baselines, sequential fine-tuning, multilingual fine-tuning, continual pre-training, and a control experiment using Turkish as an unrelated language. The strongest system, continual pre-training on Sinhala followed by fine-tuning on Dhivehi with \texttt{KenLM}, achieves 12.89\% WER and 2.70\% CER, outperforming the Dhivehi-only baseline by 13.50\% WER and 3.02\% CER. The Turkish control experiment confirms that observed improvements stem from linguistic relatedness; adaptation strategy and decoding configuration are also critical.

\end{abstract}


\section{Introduction}

Modern ASR systems achieve strong performance for high-resource languages such as English and Mandarin, which benefit from large-scale speech datasets~\cite{kahn2020librilight}. However, the majority of the world’s approximately 7,000 languages remain under-represented in speech technology, restricting access to digital services, educational tools, and communication technologies for hundreds of millions of speakers~\cite{pratap2024mms}.

Dhivehi (also called Maldivian), the national language of the Maldives~\cite{maldives_constitution_2008}, is spoken by roughly 335,000 to 410,000 people across the Maldives, Minicoy (India), and expatriate communities~\cite{gnanadesikan2017dhivehi}. Mozilla Common Voice provides only 61 recorded hours of speech, of which 37 hours are validated for Dhivehi~\cite{ardila2020commonvoice}, far below the data volumes typically required for competitive ASR systems.

Self-supervised models such as Wav2Vec~\cite{baevski2020wav2vec2} and multilingual extensions such as XLS-R~\cite{babu2022xlsr} have substantially reduced labelled data requirements through large-scale pre-training on raw audio. Fine-tuning on limited target data often underperforms because pretrained encoders lack exposure to the target language's acoustic characteristics. Cross-lingual transfer leveraging data from a related, higher-resource language offers a complementary strategy that has shown promise across several language families~\cite{conneau2021xlsr, nowakowski2023multilingual, getman2024sami}.

\begin{figure*}[t]
\centering
\includegraphics[
  width=0.8\textwidth,
  trim=4cm 5cm 3cm 3cm,
  clip
]{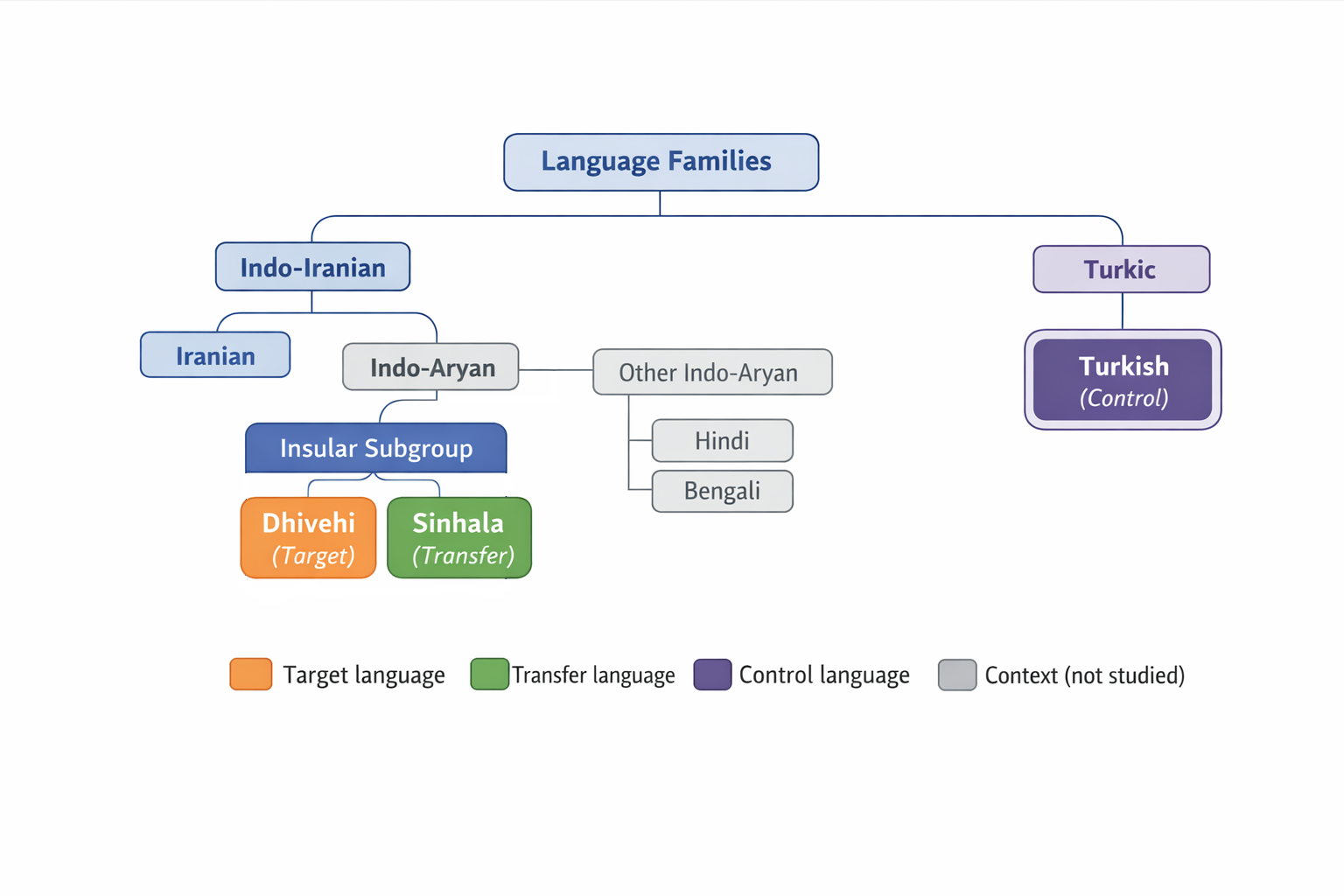}
\caption{Genealogical relationship of Sinhala, Dhivehi, and Turkish. \\
\scriptsize \textbf{Note:} Sinhala and Dhivehi belong to the Insular Indo-Aryan subgroup, while Turkish belongs to the Turkic language family and is used in this study as an unrelated control language.}
\label{fig:language-family-tree}
\end{figure*}

Dhivehi and Sinhala both belong to the Insular Indo-Aryan subgroup, sharing phonological, lexical, and structural features documented in comparative linguistic work~\cite{gnanadesikan2017dhivehi}. Figure~\ref{fig:language-family-tree} illustrates their genealogical relationship and contrasts it with Turkish, which is used in this study as a control experiment and belongs to a different language family.

Table~\ref{tab:lexical} illustrates this through sentence-level comparisons, where cognates such as \textit{magu}/\textit{maga} (``road''), \textit{gamu}/\textit{gama} (``village''), and \textit{kan}/\textit{kana} (``ear'') appear in natural utterances. These shared properties may correspond to overlapping acoustic patterns that a pretrained encoder can leverage during cross-lingual transfer. Sinhala offers approximately 220 hours of openly available transcribed speech~\cite{openslr30_sinhala_tts}, roughly four times the validated Dhivehi data, providing the necessary resource to evaluate this hypothesis.

\begin{table}[t]
\caption{Dhivehi-Sinhala sentence comparisons showing shared Indo-Aryan cognates (romanised).}
\label{tab:lexical}
\centering
\small
\resizebox{\columnwidth}{!}{
\begin{tabular}{@{}lll@{}}
\toprule
\textbf{Meaning} & \textbf{Dhivehi} & \textbf{Sinhala} \\
\midrule
I look at the sea & \textit{aharen muhudhu balanee}   & \textit{mama muhuda balanav\={a}} \\
The village is far   & \textit{gamu dhuru vey}   & \textit{gama durayi} \\
I eat fish           & \textit{aharen mas kanee} & \textit{mama mas kanav\={a}} \\
My ear hurts         & \textit{kan dhanee}       & \textit{mage kana ridenav\={a}} \\
The road is long     & \textit{magu dhigu}       & \textit{maga digayi} \\
\bottomrule
\end{tabular}}
\end{table}

Despite growing interest in multilingual ASR, research on the Sinhala and Dhivehi languages remains extremely limited. Dhivehi is not adequately covered in existing multilingual ASR frameworks. Ahmed~\cite{ahmed2023dhivehi} had to use Sinhala-language tokens when fine-tuning Whisper for Dhivehi because the model did not provide explicit support for Dhivehi. As Sinhala and Dhivehi are both Indo-Aryan languages, their shared historical and structural characteristics may provide useful conditions for related-language transfer in low-resource ASR. Current research has not systematically examined this relationship~\cite{dhasmana2026dialect, dhamecha2021role}. These gaps motivate the following research questions:

\begin{enumerate}
\item To what extent does pre-training on Sinhala speech improve Dhivehi ASR performance compared to a Dhivehi-only baseline?
\item How effectively can Sinhala and Dhivehi be represented within a shared latent space for cross-lingual ASR?
\item What role does linguistic relatedness play in cross-lingual transfer performance between Sinhala and Dhivehi?
\item How can ASR evaluation be standardised to ensure fair comparison across transfer learning strategies?
\end{enumerate}

By addressing these research questions, this study offers several contributions:

\begin{itemize}
\item Strong baselines for Sinhala and Dhivehi that exceed existing results in the literature.
\item A controlled comparison of three cross-lingual transfer paradigms for Sinhala to Dhivehi ASR, including continual pre-training (CPT), sequential fine-tuning, and multilingual fine-tuning with and without explicit language ID tokens.
\item A control experiment using Turkish speech to isolate the effect of linguistic relatedness from generic multilingual data augmentation.
\item A reproducible framework for future cross-lingual transfer learning studies.
\end{itemize}

\section{Related Work}

\subsection{Self-supervised and Multilingual Speech Models}

The transition from supervised acoustic modelling to self-supervised pre-training has significantly advanced low-resource ASR systems. Wav2Vec~\cite{baevski2020wav2vec2} learns contextualised speech
representations through masked prediction and contrastive objectives
over raw audio, enabling strong downstream ASR with
as little as ten minutes of labelled data. XLS-R~\cite{conneau2021xlsr} extended
this framework across 53 languages, while XLS-R~\cite{babu2022xlsr} scaled
to 128 languages and nearly half a million hours. MMS~\cite{pratap2024mms}
further expanded coverage to over 1,000 languages through
massive multilingual pre-training.

Whisper is a modern, large-scale, multilingual encoder-decoder model~\cite{radford2023whisper}. By relying on previously generated outputs, such models can capture language structure more naturally than pure CTC models, but they often require more data and may be less stable in very low-resource settings

\subsection{Dhivehi and Sinhala ASR}

The strongest published Dhivehi ASR work compares multilingual models based on XLS-R, MMS on Dhivehi Common Voice~13.0~\cite{ahmed2023dhivehi}. Its most important finding is that subword modelling clearly outperforms a pure character vocabulary for Dhivehi. The best system combines MMS 1B fine-tuning with a language-aware subword vocabulary, 5-gram  \texttt{KenLM} decoding, and automatic spelling correction, reducing WER to 14.26\% and improving substantially over the earlier reported Dhivehi baseline of 20.95\%~\cite{ahmed2023dhivehi}.

For Sinhala, prior work has compared hybrid deep neural network-hidden Markov model (DNN-HMM) and end-to-end lattice-free maximum mutual information (LF-MMI) systems on 40 hours of speech, reporting 27.79\% and 28.55\% WER, respectively~\cite{gamage2021improving}. More recent work shows that pretrained e2e models can achieve 23.38\% WER~\cite{gamage2024applicability}, while transfer learning with Mozilla DeepSpeech, \texttt{KenLM} decoding, and data augmentation has further reduced WER to 17.19\%~\cite{nanayakkara2024sinhala_transfer}. Taken together, these studies show that Sinhala ASR benefits from transferred representations. However, no prior work has investigated whether Sinhala can serve as a source language for targeted cross-lingual transfer to Dhivehi.

\subsection{Transfer-learning}

Yu et al.~\cite{yu2019tujiae2e} showed that Chinese-to-Tujia transfer reduced the recognition error rate to 46.19\%, improving performance by 2.11 percentage points over training on Tujia alone. Li~\cite{li2024cantonese_transfer} found that a Mandarin pretrained Wav2Vec-XLS-R model reduced Cantonese CER from about 0.30 to 0.20. Pillai et al.~\cite{pillai2024multistage} showed that Tamil to Malasar multistage transfer reduced WER to 51.9\%, and further to 47.3\% after punctuation filtering, while Kim and Kang~\cite{kim2022kwav2vec} reported over 20\% relative WER improvement by further pre-training English Wav2Vec~2.0 on Korean speech. \v{C}erniavski~\cite{cerniavski2022crosslingual} found that zero-shot transfer across Swedish, Danish, and Norwegian remained competitive, with relative degradation generally within 10--25\% compared with fully supervised systems, and multilingual decoding further reduced WER with gains of up to 15--20\% over monolingual Danish and Norwegian models. Nowakowski et al.~\cite{nowakowski2023multilingual} showed that CPT improved Sakhalin Ainu ASR from 42.3\% WER to 29.8\%, and that adding related Hokkaido Ainu yielded 30.5\% WER, whereas unrelated English performed worse at 40.1\% WER. Elamin et al.~\cite{chanie2023multilingual}, reported WER of about 18--22\% for Swahili, 20--25\% for Kinyarwanda, and 22--28\% for Luganda, with only 2--5\% relative degradation under code switching. Anidjar et al.~\cite{anidjar2023crossing} demonstrated that semantically aligned multilingual datasets improve ASR performance, for example, reducing Russian Portuguese WER to 26.22\% compared with 50.93\% under low semantic similarity conditions. Getman et al.~\cite{getman2024sami} showed that CPT and extended fine-tuning outperform direct fine-tuning for Northern S\'{a}mi, reducing XLS-R WER from 37.25\% to 30.14\% with 20 hours of data and to 28.84\% WER and 7.36\% CER.

\section{Methodology}

\subsection{Data Collection and Preprocessing}

\begin{table}[t]
\caption{Summary of speech datasets used in this study.}
\label{tab:datasets}
\centering
\small
\begin{tabular}{@{}lccc@{}}
\toprule
\textbf{Language} & \textbf{Source} & \textbf{Hours} & \textbf{Utterances} \\
\midrule
Sinhala & OpenSLR SLR52 & ${\sim}$224 & 185{,}000 \\
Dhivehi & Common Voice 22.0 & ${\sim}$61 & 15{,}000 \\
Turkish & Common Voice 22.0 & ${\sim}$60 & 55{,}000 \\
\bottomrule
\end{tabular}
\end{table}

Table~\ref{tab:datasets} summarises the datasets used in this study. For Sinhala, we use the Large Sinhala ASR training dataset from OpenSLR (SLR52), originally introduced by Kjartansson et al.~\cite{openslr30_sinhala_tts}, collected through crowdsourcing in Sri Lanka. For Dhivehi, we use the Dhivehi subset of Mozilla Common Voice~24.0~\cite{ardila2020commonvoice}. Although the full Dhivehi subset contains approximately 61\,h of audio, only the validated portion 37\,h is used for supervised training. Common Voice data is collected through a volunteer-driven process in which speakers record prompted sentences, and other contributors validate the recordings for quality and accuracy. Turkish data from Common Voice, matched in volume to the Sinhala subset, is used in the control experiment.

All audio is converted to a common format, resampled to 16\,kHz. Text transcriptions undergo Unicode normalisation (NFC), punctuation and spacing normalisation, non-linguistic symbol removal, lowercasing where needed, and language-specific script normalisation. For multilingual experiments, Sinhala is subsampled to 30\,h or 60\,h to reduce source language dominance, and \texttt{<si>} and \texttt{<dv>} tokens are prepended to each transcription.

\subsection{Model Selection and Tokenisation}

Wav2Vec was selected as the primary ASR model because it provides strong low-resource performance and a clear separation between self-supervised pre-training and supervised fine-tuning, enabling systematic comparison of transfer learning strategies. It also supports modular integration of external language models through CTC decoding. Compared with encoder-decoder models such as Whisper, it is more flexible for controlled experiments and more computationally efficient. Character-level tokenisation was used because limited text data and script differences make subword learning less reliable for Sinhala and Dhivehi.

\subsection{External Language Model}

To improve decoding quality, we incorporate an external language model during inference. Specifically, we use \texttt{KenLM}~\cite{zaiem2023finetuning}, a widely used toolkit for building efficient $N$-gram language models, together with \texttt{pyctcdecode}, a beam-search decoder designed for CTC-based speech recognition systems. A 5-gram \texttt{KenLM} model is trained on transcriptions and integrated into decoding through shallow fusion.

During inference, \texttt{pyctcdecode} combines the acoustic scores produced by the CTC acoustic model with the scores from the external language model in order to favour more plausible word sequences. The decoding score for a text hypothesis $h$ given audio input $X$ is defined as
\begin{equation}
\text{Score}(h) = \log P_{\text{CTC}}(h \mid X) + \alpha \log P_{\text{LM}}(h) + \beta |h|,
\end{equation}
where $P_{\text{CTC}}(h \mid X)$ denotes the score assigned by the acoustic model, $P_{\text{LM}}(h)$ is the language model probability, and $|h|$ is the hypothesis length. The hyperparameter $\alpha = 0.5$ controls the contribution of the language model, while $\beta = 1.0$ provides a length compensation term during decoding.

The beam width is set to 64. Empirical tuning showed that increasing the beam width to 128 degraded performance, likely because the larger search space retained more acoustically weak hypotheses during decoding.

\subsection{Experimental Design}

We conducted seventeen runs using six different experimental combinations. The experiment scripts are available in the project's GitHub\footnote{\scriptsize \url{https://github.com/lukmalilyas/From-Sinhala-to-Dhivehi-ASR}} repository.

\textbf{(1) Dhivehi only baseline:} Fine-tune the pretrained encoder on Dhivehi data only. This establishes the primary reference for all transfer comparisons.

\textbf{(2) Sinhala only baselines:} Fine-tune on 30 hours and 60 hours Sinhala subsets to validate the source language performance. The 60-hour model serves as the source checkpoint for sequential transfer.

\textbf{(3) Sequential fine-tuning:} First fine-tune on 60 hours of Sinhala, then continue fine-tuning on Dhivehi. This tests whether Sinhala provides a better initialisation for Dhivehi adaptation.

\textbf{(4) Multilingual fine-tuning:} Joint training on Sinhala + Dhivehi within a shared model, tested in four variants: with/without language ID tokens, and at 30/60 hours Sinhala proportions. This evaluates whether simultaneous cross-lingual exposure encourages beneficial shared representations.

\textbf{(5) Continual pre-training:} Continue the self-supervised pre-training objective on Sinhala audio using XLS-R, then fine-tune on Dhivehi. This targets representation-level adaptation rather than task-specific transfer.

\textbf{(6) Turkish control:} Replicate the no token multilingual setup by replacing Sinhala with an equivalent volume of Turkish data. Turkish, a Turkic language with no genealogical relationship to Dhivehi, controls for generic multilingual data augmentation effects.

\subsection{Training Configuration}

All experiments use a batch size of 16, gradient accumulation of 2 (effective batch size of 32), a learning rate of $5\times10^{5}$ with 500 warm-up steps, 10 epochs, FP16 mixed precision, and gradient checkpointing.

\subsection{Evaluation}

System performance is evaluated on Dhivehi ASR using Word Error Rate (WER) as the primary metric and Character Error Rate (CER) as a secondary metric.
\begin{equation}
\text{WER} = \frac{S + I + D}{N}
\end{equation}

WER is the primary evaluation metric because it directly reflects word-level recognition quality and is the most widely used measure in ASR benchmarking. However, WER can treat minor spelling errors and completely incorrect words equally. To provide a more fine-grained analysis, we also report CER.

\section{Experiments and Results}

\subsection{Main Results}

\begin{table*}[t]
\caption{Dhivehi target ASR results across all experimental conditions.}
\label{tab:results}
\centering
\small
\setlength{\tabcolsep}{3.5pt}
\resizebox{0.6\textwidth}{!}{%
\begin{tabular}{@{}lccccc@{}}
\toprule
\textbf{Configuration} & \textbf{LM} & \textbf{WER} & \textbf{CER} & \textbf{$\Delta$W} & \textbf{$\Delta$C} \\
\midrule
Dhivehi only baseline & \checkmark & 13.50 & 3.02 &  -  &  -  \\
Dhivehi only baseline & $\times$ & 41.27 & 6.19 &  -  &   -  \\
\midrule
Sequential (Si$\to$Dv) & \checkmark & 15.15 & 3.48 & +1.65 & +0.46 \\
Sequential (Si$\to$Dv) & $\times$ & 43.55 & 6.69 & +2.28 & +0.50 \\
\midrule
Multi 60 hrs Si + LID & \checkmark & 18.46 & 3.72 & +4.96 & +0.70 \\
Multi 60 hrs Si + LID & $\times$ & 42.29 & 6.40 & +1.02 & +0.21 \\
Multi 60 hrs Si, no LID & \checkmark & 13.26 & 3.08 & $-$0.24 & +0.06 \\
Multi 60 hrs Si, no LID & $\times$ & 42.09 & 6.30 & +0.82 & +0.11 \\
Multi 30 hrs Si, no LID & \checkmark & 13.34 & 3.04 & $-$0.16 & +0.02 \\
Multi 30 hrs Si, no LID & $\times$ & 41.94 & 6.33 & +0.67 & +0.14 \\
\midrule
Cont.\ pretrain Si$\to$Dv & \checkmark & \textbf{12.89} & \textbf{2.70} & \textbf{$-$0.61} & \textbf{$-$0.32} \\
Cont.\ pretrain Si$\to$Dv & $\times$ & 40.54 & 5.95 & $-$0.73 & $-$0.24 \\
\midrule
Turkish ctrl, no LID & \checkmark & 13.77 & 3.24 & +0.27 & +0.22 \\
Turkish ctrl, no LID & $\times$ & 43.02 & 6.60 & +1.75 & +0.41 \\
\bottomrule
\multicolumn{6}{@{}p{8.5cm}}{\footnotesize \textbf{Note:} LM = language model decoding with \texttt{KenLM}; Si = Sinhala; Dv = Dhivehi; ctrl = turkish control; LID = language identification token. $\Delta$WER and $\Delta$CER are relative to the Dhivehi-only baseline. Negative values indicate improvement.}
\end{tabular}}
\end{table*}

Table~\ref{tab:results} presents all Dhivehi target results. The best system is continual pre-training with \texttt{KenLM}, achieving \textbf{12.89\% WER} and \textbf{2.70\% CER}, the only configuration that improves over the baseline in both metrics under both decoding conditions. Multilingual fine-tuning without language ID tokens 60 hrs Sinhala achieves 13.26\% WER with  \texttt{KenLM}, modestly improving WER over the 13.50\% baseline. The 30 hrs Sinhala variant is similarly competitive, achieving 13.34\% WER. Sequential fine-tuning and multilingual training with language ID tokens both degrade performance substantially.

\subsection{Effect of  \texttt{KenLM} Decoding}

\begin{table}[t]
\caption{Absolute WER reduction from enabling  \texttt{KenLM} decoding.}
\label{tab:kenlm}
\centering
\small
\setlength{\tabcolsep}{4pt}
\begin{tabular}{@{}lccc@{}}
\toprule
\textbf{Configuration} & \textbf{No LM} & \textbf{+KenLM} & \textbf{$\Delta$WER} \\
\midrule
Dhivehi only baseline & 41.27 & 13.50 & $ $27.77 \\
Sequential (Si$\to$Dv) & 43.55 & 15.15 & $ $28.40 \\
Multi 60 hrs Si $+$ LID & 42.29 & 18.46 & $ $23.83 \\
Multi 60 hrs Si $-$ LID & 42.09 & 13.26 & $ $28.83 \\
Cont.\ pretrain Si$\to$Dv & 40.54 & 12.89 & $ $27.65 \\
Turkish ctrl $-$ LID & 43.02 & 13.77 & $ $29.25 \\
\bottomrule
\end{tabular}
\end{table}

Table~\ref{tab:kenlm} quantifies the  \texttt{KenLM} effect. Every configuration improves by 24-29 absolute WER points when  \texttt{KenLM} is enabled. Without \texttt{KenLM}, all systems cluster in a high error band around 40-44\% WER, making acoustic transfer differences nearly indistinguishable. Once \texttt{KenLM} is applied, the systems separate meaningfully, and the differences in transfer pathways become visible. This is the single largest and most consistent effect observed in the study, demonstrating that in this low-resource setting, the external language model is a dominant determinant of final transcription quality.

Reducing the beam width from 128 to 64 consistently improved performance by approximately 10 WER points across all evaluated configurations, highlighting the importance of decoder tuning.

\subsection{Role of Language ID Tokens}

\begin{table}[t]
\caption{Effect of language ID tokens in multilingual fine-tuning 60 hours Sinhala.}
\label{tab:lid}
\centering
\small
\setlength{\tabcolsep}{3pt}
\begin{tabular}{@{}cccccc@{}}
\toprule
\textbf{LID} & \textbf{LM} & \textbf{Dv-WER} & \textbf{Dv-CER} & \textbf{Si-WER} & \textbf{Si-CER} \\
\midrule
Yes & \checkmark & 18.46 & 3.72 & 11.21 & 2.56 \\
Yes & $\times$ & 42.29 & 6.40 & 23.27 & 4.44 \\
No & \checkmark & 13.26 & 3.08 & 6.49 & 1.78 \\
No & $\times$ & 42.09 & 6.30 & 23.23 & 4.47 \\
\bottomrule
\end{tabular}
\end{table}

Table~\ref{tab:lid} reveals an important finding: the multilingual model \emph{without} language ID tokens substantially outperforms the variant \emph{with} tokens for both Dhivehi 13.26\% vs.\ 18.46\% WER and Sinhala 6.49\% vs.\ 11.21\% WER when decoded with  \texttt{KenLM}. This counterintuitive result suggests that in this bilingual setting, explicit language conditioning introduces harmful decoder separation rather than beneficial disambiguation.

\subsection{Effect of Sinhala Data Proportion}

\begin{table}[t]
\caption{Effect of Sinhala data proportion on both languages in multilingual fine-tuning without language ID tokens.}
\label{tab:proportion}
\centering
\small
\setlength{\tabcolsep}{3pt}
\begin{tabular}{@{}cccccc@{}}
\toprule
\textbf{Si \%} & \textbf{LM} & \textbf{Dv-WER} & \textbf{Dv-CER} & \textbf{Si-WER} & \textbf{Si-CER} \\
\midrule
60 hrs & \checkmark & 13.26 & 3.08 & 6.49 & 1.78 \\
30 hrs & \checkmark & 13.34 & 3.04 & 8.35 & 2.26 \\
60 hrs & $\times$ & 42.09 & 6.30 & 23.23 & 4.47 \\
30 hrs & $\times$ & 41.94 & 6.33 & 26.68 & 5.22 \\
\bottomrule
\end{tabular}
\end{table}

Table~\ref{tab:proportion} shows that Dhivehi performance is remarkably stable across the two Sinhala proportions $\Delta$WER $= 0.08$ with  \texttt{KenLM} in multilingual fine-tuning, while Sinhala baseline performance degrades when its training share is halved 6.49\%$\to$8.35\% WER. This difference indicates that the multilingual benefit to the target language saturates at a relatively low level of source-language volume, suggesting that even modest amounts of related language data may suffice.

\subsection{Unrelated Language Control Experiment}

The Turkish control provides critical evidence for interpreting the multilingual results. Replacing Sinhala with Turkish degrades performance relative to both the Sinhala paired model (13.77\% vs 13.26\% with \texttt{KenLM}) and the Dhivehi-only baseline (13.50\%). Without  \texttt{KenLM}, the gap widens further 43.02\% vs 42.09\% vs 41.27\%. This confirms that the modest multilingual improvements from Sinhala reflect genuine phonological and acoustic similarities between the two Insular Indo-Aryan languages, rather than a generic data-augmentation effect.

\subsection{Sinhala Baseline Validation}

\begin{table}[t]
\caption{Sinhala only baseline results validating the source language training pipeline.}
\label{tab:sinhala}
\centering
\small
\begin{tabular}{@{}lcc@{}}
\toprule
\textbf{Configuration} & \textbf{WER (\%)} & \textbf{CER (\%)} \\
\midrule
30 hrs Sinhala, no KenLM & 26.33 & 5.13 \\
60 hrs Sinhala, no KenLM & 23.83 & 4.56 \\
60 hrs Sinhala + KenLM & 7.00 & 1.89 \\
\bottomrule
\end{tabular}
\end{table}

Table~\ref{tab:sinhala} shows that the Sinhala source setup is reliable, achieving 7.00\% WER with  \texttt{KenLM}. This means the weaker sequential and multilingual transfer results are more likely due to limitations in the transfer approach, rather than a weak Sinhala source model.

\subsection{Comparison with Published Benchmarks}

\begin{table}[t]
\caption{Comparison with Published Dhivehi and Sinhala ASR Benchmarks.}
\label{tab:comparison}
\centering
\small
\resizebox{0.95\columnwidth}{!}{
\begin{tabular}{lcc}
\toprule
\textbf{System} & \textbf{WER (\%)} & \textbf{Note} \\
\midrule
\multicolumn{3}{c}{\textit{Dhivehi}} \\\hline
Earlier baseline (lit.) & 20.95 & Historical ref. \\
Ahmed~\cite{ahmed2023dhivehi} best & 14.26 & MMS 1B + LM \\
This work: Dv only + LM & 13.50 & Wav2Vec2-BERT \\
This work: Multi $ $LID + LM & 13.26 & 40\% Sinhala \\
This work: CPT + LM & \textbf{12.89} & Best result \\
\midrule
\multicolumn{3}{c}{\textit{Sinhala}} \\\hline
Nanayakkara~\cite{nanayakkara2024sinhala_transfer} & 17.19 & DeepSpeech + aug. \\
This work: 40\% Si + LM & 7.00 & Wav2Vec2-BERT \\
\bottomrule
\multicolumn{3}{@{}p{8.5cm}}{\footnotesize \textbf{Note:} Direct comparison requires caution due to differences in data version, preprocessing, and tokenisation.}
\end{tabular}}
\end{table}

Table~\ref{tab:comparison} compares our results with existing literature. Even the Dhivehi-only baseline 13.50\% numerically surpasses the strongest published Dhivehi benchmark 14.26\%~\cite{ahmed2023dhivehi}. The best transfer system, 12.89\%, further improves upon this already competitive baseline. The Sinhala result 7.00\% is substantially lower than the literature benchmark 17.19\%, though direct comparison is limited by differences in evaluation conditions. Note that Ahmed~\cite{ahmed2023dhivehi} evaluated models using Common Voice 13.0, whereas this work uses Common Voice 22.0 (we use only the validated Dhivehi subset for supervised training);

\section{Discussion}

\subsection{Transfer Strategy Matters More Than Relatedness Alone}
The main finding is a qualified one: Sinhala improves Dhivehi ASR, but only through certain transfer strategies. Continual pre-training is the strongest approach because it adapts the encoder's speech representations before introducing Dhivehi-specific supervision, thereby retaining reusable acoustic features. Multilingual fine-tuning without language ID tokens also gives a small gain. By contrast, sequential fine-tuning does not outperform the Dhivehi-only baseline, suggesting that stronger adaptation to the Sinhala supervised task may reduce the model's transfer to Dhivehi.

With \texttt{KenLM}, the Dhivehi-only baseline achieves 13.50\% WER, already numerically better than the strongest Dhivehi benchmark reported in the literature review; the transfer gains are therefore meaningful because they are achieved over a strong target-only baseline.

\subsection{Language ID Tokens Are Not Universally Beneficial}
The multilingual experiments show that shared modelling can help with Dhivehi, but the conditioning method matters. In this low-resource, bilingual, character-level setting, multilingual fine-tuning without language ID tokens performs much better than the version with them. This suggests that useful sharing is better supported when the model is allowed to learn common structure without explicit prefix-level separation. The weaker token-based result does not mean that language tokens are always harmful, but it does show that, in this setting, extra language conditioning may impose more separation than the learning problem can usefully support.

\subsection{Decoding Support Is a First Order Effect}
The effect of  \texttt{KenLM} is too large to be treated as a minor implementation detail. In the Dhivehi-only system, WER falls from 41.27\% to 13.50\%, and similarly large reductions appear across the transfer settings, including the Turkish control. This shows that the best systems in the study should be understood not only as transfer learning results, but as transfer plus decoding results. Strong acoustic representations alone are not enough to explain the final performance. For low-resource Dhivehi ASR, decoder support is therefore essential, both for fair interpretation of model comparisons and for practical system design.

\subsection{Linguistic Relatedness Alone Is Not Enough}
The Turkish control provides the clearest evidence that relatedness matters. Under the same multilingual setup, Sinhala improves Dhivehi WER to 13.26\% with  \texttt{KenLM}, while Turkish reaches 13.77\%, compared with 13.50\% for the Dhivehi-only baseline. If the effect were only data augmentation, Turkish should have performed similarly. Instead, the Sinhala advantage suggests that phonological and acoustic overlap between the two related languages supports useful transfer. However, relatedness alone is not enough as sequential transfer still fails to beat the baseline. The most appropriate conclusion is therefore that relatedness helps, but the training pipeline determines whether that benefit is realised.

\subsection{Limitations}

Computational constraints limited the overall training scale, the number of repeated runs, and the extent of hyperparameter tuning; consequently, we did not evaluate models across multiple data splits or perform cross-validation. Continual pre-training used the XLS-R architecture because Wav2Vec2-BERT did not support the same CPT setup, creating an architectural inconsistency across experiments. Some pipeline instability affected intermediate estimates when integrating the pyctcdecode library, but final reported evaluations were manually verified.

\section{Conclusion}

We presented a controlled evaluation of Sinhala-to-Dhivehi cross-lingual transfer for low-resource ASR across five transfer paradigms and 17 combinations. Continual pre-training on Sinhala followed by fine-tuning to Dhivehi with \texttt{KenLM} achieves the best performance (12.89\% WER and 2.70\% CER), while multilingual fine-tuning without language ID tokens provides a competitive alternative. Sequential transfer and multilingual training with language tokens do not improve over the strong Dhivehi-only baseline. The Turkish control experiment confirms that observed gains can be attributed to linguistic relatedness. However, \texttt{KenLM} decoding is the single most impactful component, contributing far more than any transfer strategy.

These results show three key implications for low-resource ASR research. First, language-relatedness is necessary, but the transfer learning strategy determines whether shared linguistic features lead to measurable performance improvement. Second, design assumptions from large-scale multilingual systems, such as the benefit of language ID tokens, do not necessarily transfer to low-resource bilingual settings and must be validated empirically. Third, external language modelling remains essential and should not be treated as a secondary concern in the ASR pipeline.

Several directions for future research follow from this study. First, qualitative error analysis should examine which Dhivehi phonemes, word types, or linguistic structures benefit most from Sinhala-based transfer. Second, evaluation should be extended beyond the current benchmark to include a wider range of speakers, accents, domains, and recording conditions. Third, future work could compare Sinhala transfer with other related and unrelated source languages to better measure the value of linguistic relatedness. Fourth, stronger neural language models, especially Transformer-based approaches, could be explored for post-processing rather than relying solely on n-gram methods. Overall, this study provides a stronger benchmark and a reusable framework for future low-resource ASR research

% \section*{Acknowledgment}
% The authors would like to thank Google for providing the computational resources used in this study.

{\footnotesize
\bibliographystyle{IEEEtran}
\bibliography{references}
}

\end{document}